\documentclass{article}

\usepackage[preprint]{neurips_2021}

\usepackage{xcolor}         
\definecolor{linkColor}{rgb}{0.18,0.39,0.62}
\usepackage[utf8]{inputenc} 
\usepackage[T1]{fontenc}    
\usepackage[colorlinks=true,linkcolor=linkColor,citecolor=linkColor,filecolor=linkColor,urlcolor=linkColor]{hyperref}       
\usepackage{url}            
\usepackage{booktabs}       
\usepackage{amsfonts}       
\usepackage{nicefrac}       
\usepackage{microtype}      
\usepackage[most]{tcolorbox}
\usepackage{enumitem}

\usepackage{amsmath}
\usepackage{amssymb}
\usepackage{mathtools}
\usepackage{amsthm}
\usepackage{graphicx}
\usepackage{subfigure}
\usepackage{xspace}
\usepackage{pifont}

\theoremstyle{plain}
\newtheorem{theorem}{Theorem}[section]

\theoremstyle{definition}

\theoremstyle{remark}

\usepackage[capitalize,noabbrev]{cleveref}
\usepackage{framed}
\usepackage{multirow}
\usepackage{svg}
\usepackage{listings}
\usepackage{wrapfig}

\newcommand\our{\textsc{Magneto}}

\newcommand\postln{Post-LN}
\newcommand\preln{Pre-LN}
\newcommand\subln{Sub-LN}

\newcommand{\eg}{\textit{e}.\textit{g}.}
\newcommand{\xmark}{\ding{55}\xspace}%

\title{Foundation Transformers}

\author{
\vspace{-0.15in}
\\
Hongyu Wang\thanks{~Equal contribution. $\dagger$ Corresponding author.},~\ Shuming Ma\footnotemark[1],~\ Shaohan Huang,~Li Dong,~Wenhui Wang,~Zhiliang Peng,~Yu Wu \\
{Payal Bajaj,}~{Saksham Singhal,}~{Alon Benhaim,}~{Barun Patra,}~{Zhun Liu},~{Vishrav Chaudhary} \\
{Xia Song,}~\ {Furu Wei}$^\dagger$ \\
Microsoft\\
{\url{https://github.com/microsoft/unilm}}
}

\begin{document}

\maketitle
\vspace{-0.5cm}
\begin{abstract}
A big convergence of model architectures across language, vision, speech, and multimodal is emerging. However, under the same name ``Transformers'', the above areas use different implementations for better performance, e.g., Post-LayerNorm for BERT, and Pre-LayerNorm for GPT and vision Transformers.
We call for the development of \textbf{Foundation Transformer} for \textit{true general-purpose modeling}, which serves as a go-to architecture for various tasks and modalities with guaranteed training stability.
In this work, we introduce a Transformer variant, named \textbf{\textsc{Magneto}}, to fulfill the goal.
Specifically, we propose Sub-LayerNorm for good expressivity, and the initialization strategy theoretically derived from DeepNet~\citep{deepnet} for stable scaling up.
Extensive experiments demonstrate its superior performance and better stability than the de facto Transformer variants designed for various applications, including language modeling (i.e., BERT, and GPT), machine translation, vision pretraining (i.e., BEiT), speech recognition, and multimodal pretraining (i.e., BEiT-3).
\end{abstract}


\begin{table}[h]
\begin{center}
\vspace{-0.4cm}
\begin{tabular}{ll|l|c|c}
\toprule
 \multicolumn{3}{c|}{\textbf{Models}} & \bf Previous & \textbf{This work} \\
\midrule
\bf Vision & Encoder & ViT/BEiT & \preln{} & \multirow{7}{*}{\textbf{\subln{}}} \\
\cmidrule(lr){1-4}
\multirow{3}{*}{\textbf{Language}} & Encoder & BERT & \postln{} &  \\
& Decoder & GPT & \preln{} & \\
& Encoder-Decoder & NMT/BART & \postln{} & \\
\cmidrule(lr){1-4}
\bf Speech & Encoder & T-T & \preln{} & \\
\cmidrule(lr){1-4}
\bf Multimodal & Encoder & BEiT-3 & \preln{} & \\
\bottomrule
\end{tabular}
\label{tab:arch}
\end{center}
\end{table}

\vspace{-0.15cm}
\begin{figure}[h]
\centering
\vspace{-0.8cm}
\subfigure[Post-LN]{
\includegraphics[width=0.2\columnwidth]{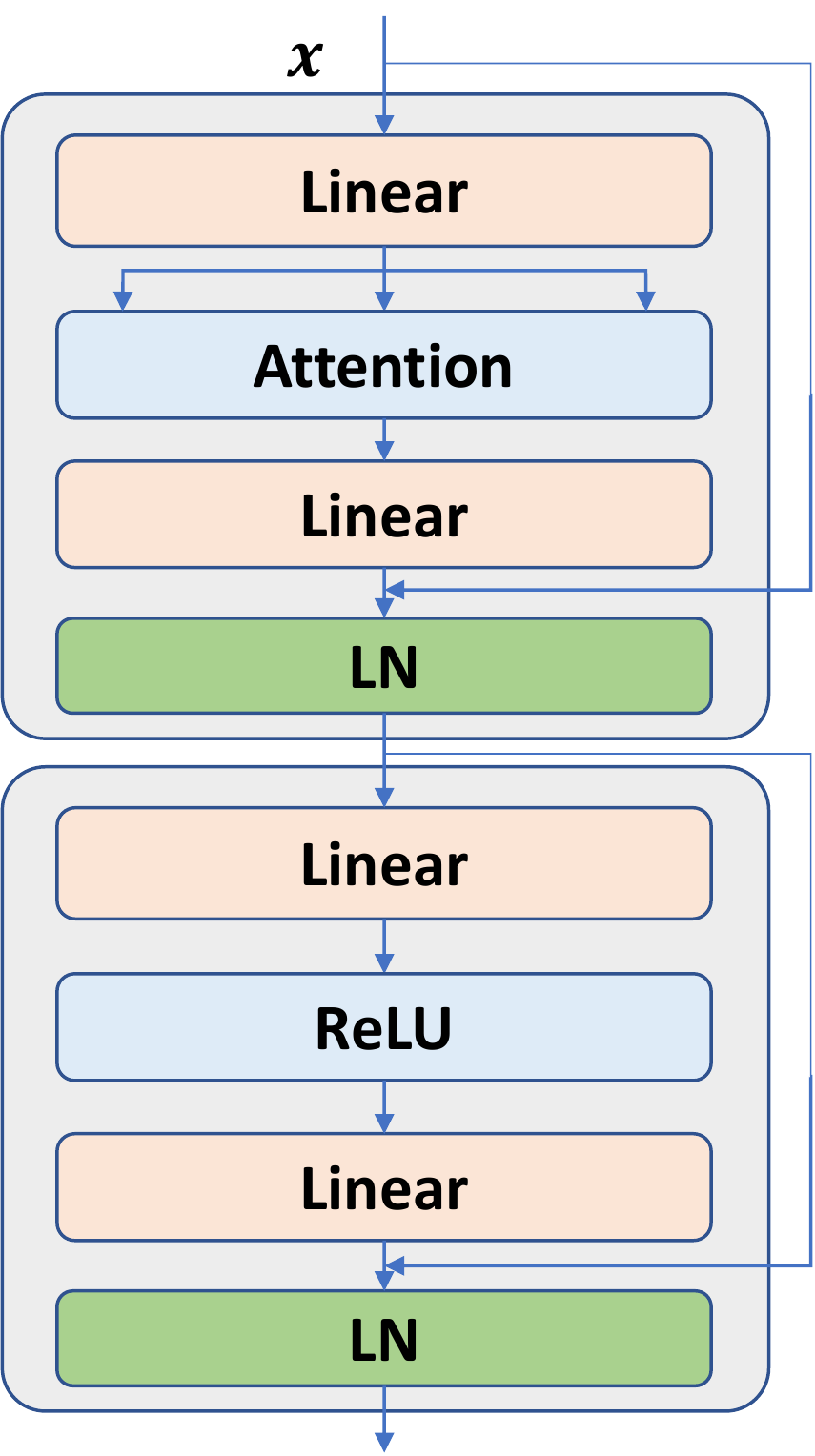}
}
\hspace{1.5cm}
\subfigure[Pre-LN]{
\includegraphics[width=0.2\columnwidth]{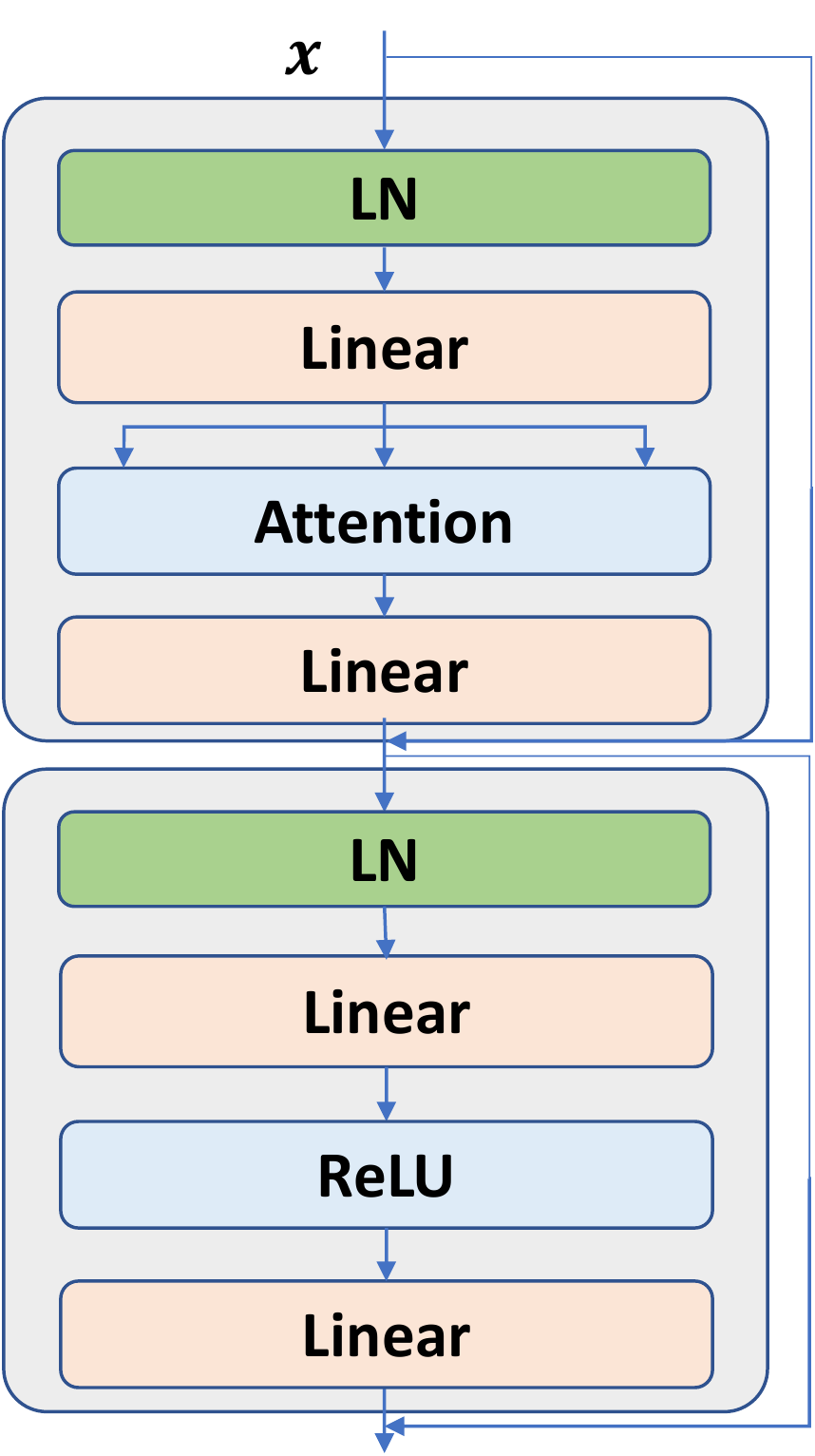}
}
\hspace{1.5cm}
\subfigure[Sub-LN]{
\includegraphics[width=0.3\columnwidth]{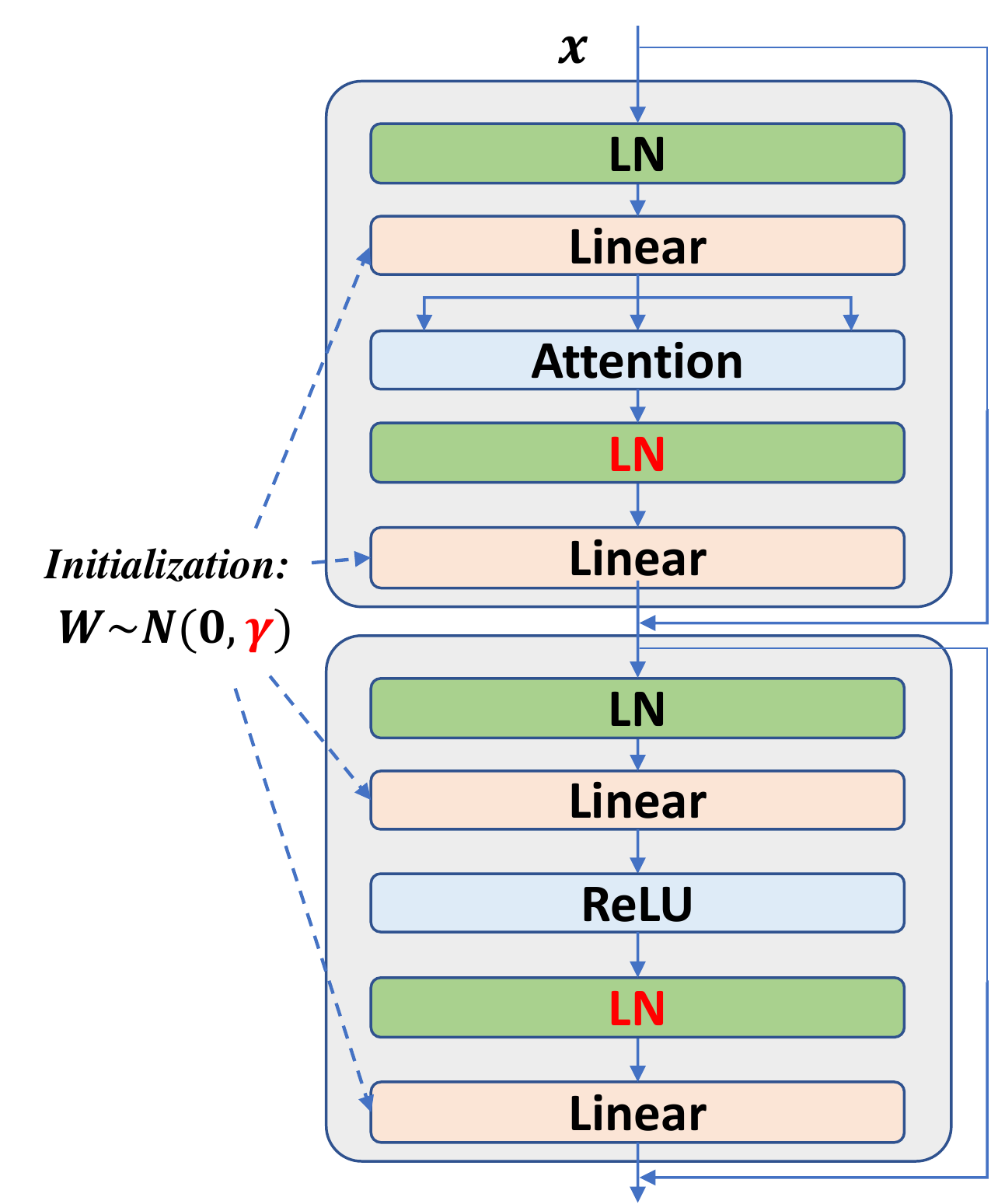}
}

\vspace{-0.2cm}
\caption{
\textbf{Top}: the architectures of SOTA models across language, vision, speech, and multimodal. \textbf{Bottom}: the proposed Foundation Transformer uses \subln{} and theoretically derived initialization.
}
\vspace{-0.3cm}
\label{fig:subln}
\end{figure}

\newpage
\section{Introduction}
\label{sec:intro}

Recent years have witnessed a big convergence of model architectures across language, vision, speech, and multimodal.
Specifically, starting from the natural language processing, Transformers~\citep{transformer} have become the de facto standard for various areas, including computer vision~\citep{vit}, speech~\citep{zhang2020transformer}, and multimodal~\citep{vilt,beit3}.
Transformers fully leverage the parallelism advantage of GPU hardware and large-scale data.
It is appealing that we can use the same network architecture for a broad range of applications. So the pretrained models can be seamlessly reused with the shared implementation and hardware optimization.
Moreover, general-purpose modeling is important to multimodal models, as different modalities can be jointly encoded and fused by one model.

However, despite using the same name ``Transformers'', there are significant differences in the implementation of the architectures for different tasks.
\cref{fig:subln} summarizes the architectures for state-of-the-art models that are widely used in various communities.
For instance, some models (e.g., GPT, and ViT) adopt Pre-LayerNorm (Pre-LN) Transformers, while others use Post-LayerNorm (Post-LN) variants (e.g., BERT, and machine translation) for better performance.
Rather than directly using the same architecture, we need to compare two Transformer variants on the specific tasks or modalities to determine the backbone, which is ineffective for model development.
More importantly, considering multimodal models, the optimal Transformer variants are usually different for input modalities.
For the example of BEiT-3~\citep{beit3} vision-language pretraining, using \postln{} is sub-optimal for vision encoding while \preln{} is sub-optimal for the language part. The true convergence of multimodal pretraining requires a unified architecture that performs well across tasks and modalities.
In addition, a pain point of Transformer architectures is training stability, especially for large-scale models.
We usually need significant efforts to tune hyperparameters or babysit training processes.

As a result, we call for developing \textbf{Foundation Transformers} for \textit{true general-purpose modeling}.
First, the desired modeling should be able to serve as a go-to architecture for various tasks and modalities, so that we can use the same backbone without trial and error.
The general-purpose design principle also greatly supports the development of multimodal foundation models, as we can use one unified Transformer for various modalities without performance degradation.
Second, the architectures should provide guaranteed training stability. The favored property can significantly mitigate the difficulty of large-scale pretraining of foundation models.

In this work, we introduce \our{} as an implementation of Foundation Transformers to fulfill the above goals.
Specifically, we introduce Sub-LayerNorm (\subln{}), which adds an extra LayerNorm to each sublayer (i.e., multi-head self-attention, and feed-forward network).
Moreover, \our{} has a novel initialization method that has a theoretical guarantee to fundamentally improve the training stability.
This allows the models to be scaled up without pain.
We evaluate \our{} on extensive tasks and modalities, namely, masked language modeling (i.e., BERT), causal language modeling (i.e., GPT), machine translation, masked image modeling (i.e., BEiT), speech recognition, and vision-language pretraining (i.e., BEiT-3).
Experimental results show that \our{} significantly outperforms de facto Transformer variants on the downstream tasks.
In addition, \our{} is more stable in terms of optimization, which allows larger learning rates to improve results without training divergence.

\section{TL;DR for Practitioners}

\begin{figure}[htbp]
\centering
\begin{minipage}[t]{0.44\columnwidth}
\vspace{0pt}
\begin{lstlisting}[language=python,mathescape]
  def subln(x):
      return x + fout(<@\textcolor{red}{LN}@>(fin(LN(x))))

  def subln_init(w):
      if w is ['ffn', 'v_proj', 'out_proj']:
          nn.init.xavier_normal_(w, gain=<@\textcolor{red}{$\gamma$}@>)
      elif w is ['q_proj', 'k_proj']:
          nn.init.xavier_normal_(w, gain=1)
\end{lstlisting}
\end{minipage}
\hfill
\begin{minipage}[t]{0.545\columnwidth}
\vspace{0pt}
\resizebox{\columnwidth}{!}{
\begin{tabular}{l|c|c}
\toprule
\multirow{2}{*}{\textbf{Architectures}} & \multicolumn{1}{c|}{\textbf{Encoder}} & \multicolumn{1}{c}{\textbf{Decoder}} \\
&  $\gamma$ &  $\gamma$ \\
\midrule
\text{Encoder-only} & \multirow{2}{*}{$\sqrt{\log{2N}}$} & \multirow{2}{*}{-}  \\
\text{(e.g., BERT, ViT)} & & \\
\text{Decoder-only} &  \multirow{2}{*}{-}  & \multirow{2}{*}{$\sqrt{\log{2M}}$}  \\
\text{(e.g., GPT)} & & \\
\text{Encoder-decoder} &  \multirow{2}{*}{$\sqrt{\frac{1}{3}\log{3M}\log{2N}}$}  &  \multirow{2}{*}{$\sqrt{\log{3M}}$} \\
\text{(e.g., NMT, BART)} & & \\
\bottomrule
\end{tabular}
}
\end{minipage}
\subfigure[Encoder or Decoder]{
\includegraphics[width=0.4\columnwidth]{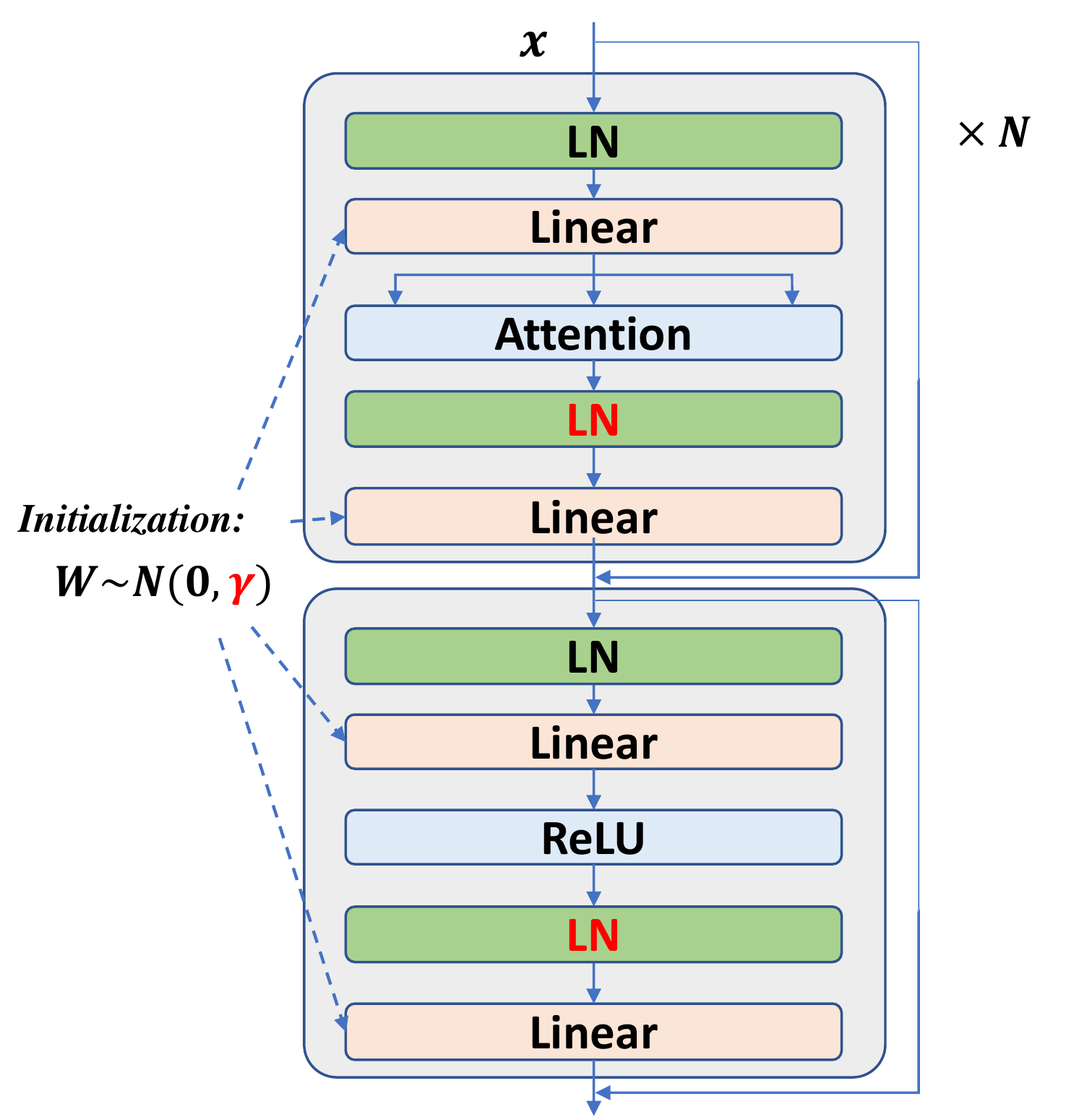}
}
\subfigure[Encoder-Decoder]{
\includegraphics[width=0.55\columnwidth]{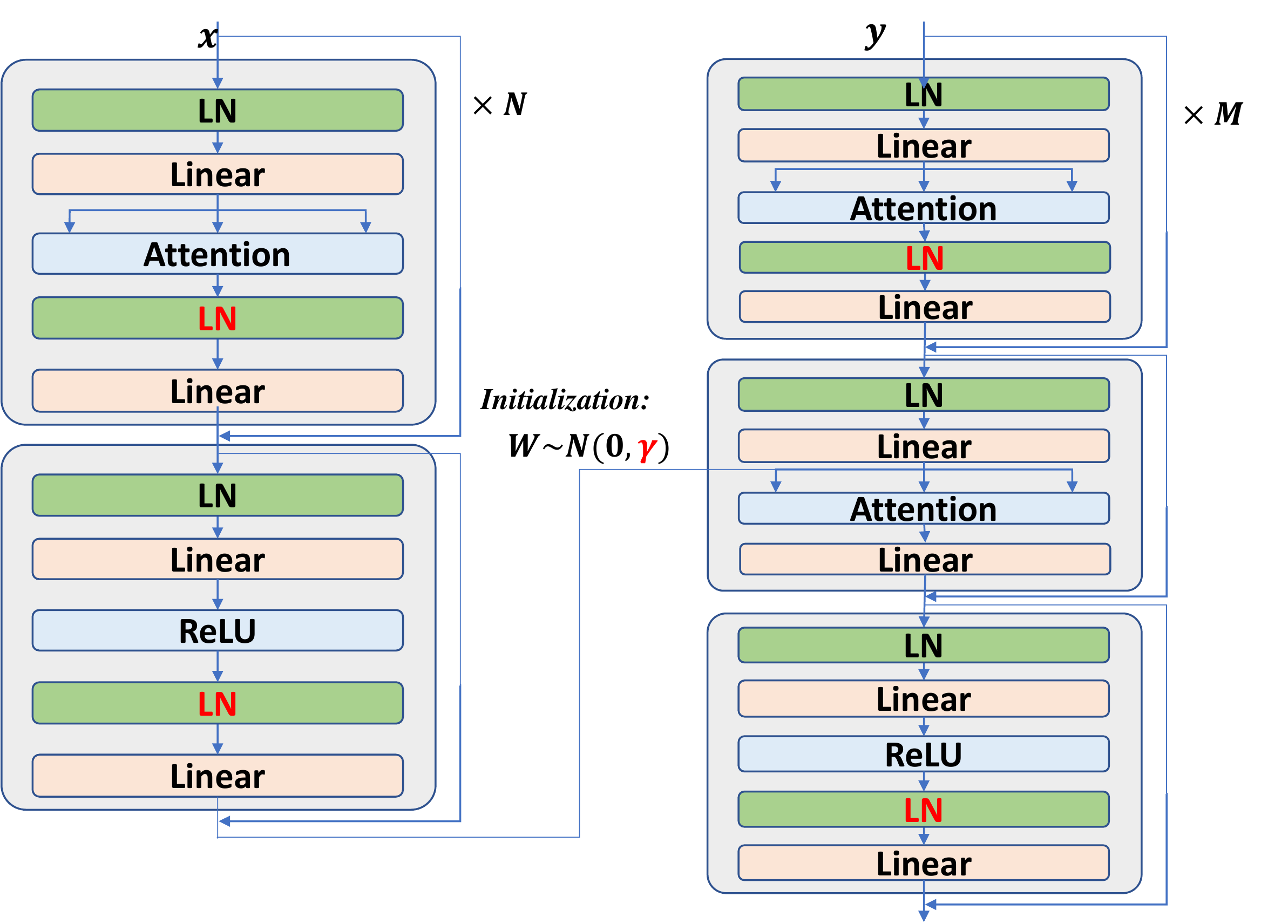}
}
\caption{\textbf{Top left}: pseudocode of \subln{}. We take Xavier initialization~\citep{xavier} as an example, and it can be replaced with other standard initialization. Notice that $\gamma$ is a constant. \textbf{Top right}: parameters of \subln{} for different architectures ($N$-layer encoder, $M$-layer decoder). \textbf{Bottom}: the layout of \subln{} for different architectures.}
\label{implementation}
\end{figure}

\cref{fig:subln} illustrates the overview of the \our{} architecture.
There are two key improvements in terms of modeling.
First, compared to the \preln{} variant, \subln{} introduces another LayerNorm inside each sublayer (i.e., multi-head self-attention, and feed-forward network): one before the input projection, and the other before the output projection.
Second, we use the initialization with the theoretical derivation from DeepNet~\citep{deepnet}, which fundamentally improves the training stability, allowing the model to be scaled up to massive sizes without pain.

As shown in \cref{implementation}, we present the implementation of \our{}.
There are only lines of code changes on top of the vanilla Transformer architecture. Notably, following the derivation from DeepNet, the weights of query projection and key projection are not scaled during initialization. Besides, there is only one LayerNorm inside the cross-attention for the encoder-decoder architecture and we do not scale the initialized weights of cross-attention.

\section{\our{}: A Foundation Transformer}
\label{sec:method}

\subsection{Architecture: Sub-LayerNorm}

Vanilla Transformers are based on either Pre-LayerNorm (\preln{}) structures or Post-LayerNorm (\postln{}). Different from them, \our{} is built on the Sub-LayerNorm (\subln{}). It inherits the multihead attentions and the feed-forward network from Transformers and introduces two layer normalization modules inside each sublayer (except the cross-attention). 

For the multihead attentions, the layer normalization modules are before the $qkv$ projection and the output projection, which can be formulated as:

\begin{align}
    Q, K, V &= W^{Q}\text{LN}(x), W^{K}\text{LN}(x), W^{V}\text{LN}(x) \\
    \text{MSA}(x) &= x + W^{O} \text{LN}(\text{Attention}(Q,K,V))
\end{align}

where $W^{Q}$, $W^{K}$, $W^{V}$, and $W^{O}$ are the parameters of the multihead self-attention.
Similarly, for the feed-forward network, the layer normalization modules are before the input projection and the output projection, which are written as:

\begin{align}
    \text{FC}_1(x) &= W^{1} \text{LN}(x) \\
    \text{FC}_2(x) &= W^{2} \text{LN}(x) \\
    \text{FFN}(x) &= \text{FC}_2(\phi(\text{FC}_1(x)))
\end{align}

where $W^{1}$ and $W^{2}$ are parameters of the feed-forward layers, and $\phi$ is the non-linear activation function.

\subsection{Initialization: Theoretical Derivation from DeepNet}

We adopt the theoretical derivation from DeepNet~\citep{deepnet} to improve the training stability.
DeepNet estimates the expected model update for \postln{} and introduces DeepNorm to bound the model update to a constant. Following DeepNet, we first estimate the expected model update of \subln{} and then demonstrate how to bound the model update with a proper initialization.

\paragraph{Expected Model Update for \preln{}}

We start with the expected model update for \preln{}.
The forward propagation for an $N$-layer \preln{} Transformer with $N$ attention sub-layers and $N$ feed-forward sub-layers can be formulated as:

\begin{align}
    &F(x; \theta) = W^{vocab} x^{e} \\
    &x^{e} = \text{LN}(x + \sum_{l=1}^{L} G^l(x^{l-1}, \theta_{el})), 
    \quad x^l = G^l(x^{l-1}, \theta_{el})
    \,\text{and}\,x^0 = x
\end{align}

where $x^{l-1}$, $x^{l}$ denotes the input and output for the $l$-th sub-layer $G^l$. If $l$ is odd, $G^l$ refers to self-attention MSA; if $l$ is even, $G^l$ refers to FFN. $x^e$ is the output of the backbone. $\theta$ denotes the parameters of output projection $W^{vocab}$ and the backbone $\{\theta_{el}\}_{l=1}^L$. $W^{vocab} \in R^{V \times d}$, where $d$ is hidden dimension, $V$ is dictionary size. $L$ equals to $2N$ for simplicity. Without the loss of generality, we set the intermediate dimension of feed-forward layers equals to hidden dimension.

Following \cite{deepnet}, the magnitude of attention output only depends on value and output projection: $\text{MSA}(X) \overset{\Theta}{=} W^{O} W^{V}\text{LN}(X)$. Similarly we have $\text{FFN}(x) = W^2\phi(W^1 \text{LN}(X))$. Therefore, for vanilla \preln{}, the forward computation of the $l$-th sub-layer can be formulated as:

\begin{equation}
    x^l = x^{l-1} + W^{l,2}\phi(W^{l,1}\text{LN}(x^{l-1}))
\end{equation}

 We introduce two constants $v_l, w_l$ to represent the scales of $W^{l,2}, W^{l,1}$ respectively. For example, the $i$-th row, $j$-th column entry of $W^{l,2}$ satisfies that:

 \begin{equation}
     W_{ij}^{l, 2} \backsim \mathcal{N}(0, \cfrac{v_l^2}{d})
 \end{equation}

We define the model update $\Delta F = || \gamma^T (F(x; \theta^*) - F(x; \theta))||$, where $\gamma, F(x) \in$ $R^{V \times 1}$. $x$ and $F(x)$ denote the input and output of the model respectively. $\gamma$ is the label of $x$, which is a one-hot vector with a single entry as 1 and all the others as 0. With above analysis, we have the following theorem to characterize $\Delta F^{pre}$ for an $N$-layer, encoder-only \preln{} Transformer under SGD update.

\begin{theorem}
\label{thm:pre_enc}
Given an $N$-layer \preln{} Transformer $F(x, \mathbf{\theta})$, the $l$-th sub-layer is formulated as $x^l = x^{l-1} + W^{l,2}\phi(W^{l,1}\text{LN}(x^{l-1}))$. Under SGD update, $\Delta F^{pre}$ satisfies: 
\begin{equation}
\label{eq:pre_bound}
    \Delta F^{pre}
    \le \eta d(\cfrac{\sum_{l=1}^L v_l^2 + w_l^2}{\sum_{n=1}^{L} v_n^2 w_n^2} + \sum_{l=1}^L \sum_{k=2}^L\cfrac{v_l^2 + w_l^2}{\sum_{n=1}^{L} v_n^2 w_n^2}\,\cfrac{v_k^2 w_k^2}{\sum_{n=1}^{k-1} v_n^2 w_n^2}))
\end{equation}
where $\eta$ is learning rate, $L$ equals to $2N$.
\end{theorem}

Based on \cref{thm:pre_enc}, with $v_l = w_l = 1$ (i.e., standard initialization) for vanilla \preln{}, we have $\Delta F^{pre} = \mathcal{O}(\eta d \log L)$, which shows that the magnitude of the model update grows logarithmically as the depth increases. It is also verified by \citet{LiyuanLiu2020UnderstandingTD}. \citet{deepnet} proves that under SGD update, the model update of vanilla \postln{} $\Delta F^{post}$ is $\mathcal{O}(\sum_{l=1}^L v_l^2 + w_l^2)$. $\Delta F^{pre}$ is much smaller than $\Delta F^{post}$ with the same model depth $L$. It indicates that the loss landscape of vanilla \preln{} is smoother than that of vanilla \postln{}, which leads to faster and more stable optimization. 

\paragraph{Expected Model Update for \our{}}

Based on the analysis on \preln{}, we further estimate the expected model update of \subln{}. With \subln{}, the forward signal propagation of the $l$-th sub-layer can be formulated as:

\begin{equation}
    x^l = x^{l-1} + W^{l,2}\text{LN}(\phi(W^{l,1}\text{LN}(x^{l-1})))
\end{equation}

We then give the expected bound of the model update's magnitude $\Delta F^{sub}$ for an $N$-layer, encoder-only \our{}. 

\begin{theorem}
\label{thm:deep_enc}
Given an $N$-layer \our{} $F(x, \mathbf{\theta})$, the $l$-th sub-layer is formulated as $x^l = x^{l-1} + W^{l,2}\text{LN}(\phi(W^{l,1}\text{LN}(x^{l-1})))$. Under SGD update, $\Delta F^{sub}$ satisfies: 
\begin{equation}
\label{eq:deep_enc_bound}
    \Delta F^{sub}
    \le \eta d(\cfrac{\sum_{l=1}^L(1 + \cfrac{v_l^2}{w_l^2})}{\sum_{n=1}^{L} v_n^2} +
    \sum_{l=1}^L\sum_{k=2}^L \cfrac{1 + \cfrac{v_l^2}{w_l^2}}{\sum_{n=1}^{L} v_n^2} \cfrac{v_k^2}{\sum_{n=1}^{k-1} v_n^2})
\end{equation}
where $\eta$ is learning rate, $L$ equals to $2N$.
\end{theorem}

When the activation of the $l$-th sub-layer explodes, it leads to $w_l \gg w_i,\, i \ne l$. \cref{eq:cmp_update} proves that the model update of \our{} is smaller than that of vanilla \preln{} in this case.

\begin{equation}
\label{eq:cmp_update}
    \cfrac{1 + \cfrac{v_l^2}{w_l^2}}{\sum_{n=1}^{L} v_n^2} = \cfrac{v_l^2 + w_l^2}{w_l^2 \sum_{n=1}^L v_n^2} \le \cfrac{v_l^2 + w_l^2}{ \sum_{n=1}^L v_n^2 w_n^2},\quad w_l \gg w_i,\, i \ne l
\end{equation}

Furthermore, we study the magnitude of model update for \our{} with the encoder-decoder architecture. $\theta_{e}$ follows the same definition as in \cref{thm:deep_enc}. Similarly $\theta_{d}$ denotes parameters of decoder. \cref{thm: deep_en2de} shows that the bound of the magnitude of model update under SGD update $\Delta F_{ed} = ||\gamma^T(F_{ed}(x, y, \theta_e^*, \theta_d^*) - F_{ed}(x, y, \theta_e, \theta_d))||$, where $x$ and $y$ denote the input of encoder and decoder respectively.

\begin{theorem}
\label{thm: deep_en2de}
Given an encoder-decoder \our{} $F_{ed}(x, y, \theta_e, \theta_d)$ with N encoder layers and M decoder layers, where the $l$-th sub-layer is formulated as $x^l = x^{l-1} + W^{l,2}\text{LN}(\phi(W^{l,1}\text{LN}(x^{l-1})))$. Under SGD update, $\Delta F_{ed}$ satisfies:
\begin{align}
\label{eq:deep_enc_dec_bound}
    &\Delta F_{ed} \le \Delta F_{d} + \sum_{l=1, l\%3=1}^{L_d} \cfrac{v_{dl}^2}{\sum_{n=1}^{L_d} v_{dn}^2}(1 + \sum_{k=2}^{L_d} \cfrac{v_{dk}^2}{\sum_{n=1}^{k-1} v_{dn}^2}) \Delta F_{e} \\
    &\Delta F_{d} \overset{\Theta}{=} \eta d(\cfrac{\sum_{l=1}^{L_d}(1 +  \cfrac{v_{dl}^2}{w_{dl}^2})}{\sum_{n=1}^{L_d} v_{dn}^2} +
    \cfrac{1}{\sum_{n=1}^{L_d} v_{dn}^2} \sum_{l=1}^{L_d}\sum_{k=2}^{L_d} (1 + \cfrac{v_{dl}^2}{w_{dl}^2}) \cfrac{v_{dk}^2}{\sum_{n=1}^{k-1} v_{dn}^2}) \\
    &\Delta F_{e} \overset{\Theta}{=} \eta d(\cfrac{\sum_{l=1}^{L_e}(1 + \cfrac{v_{el}^2}{w_{el}^2})}{\sum_{n=1}^{L_e} v_{en}^2} +
    \cfrac{1}{\sum_{n=1}^{L_e} v_{en}^2} \sum_{l=1}^{L_e}\sum_{k=2}^{L_e} (1 + \cfrac{v_{el}^2}{w_{el}^2}) \cfrac{v_{ek}^2}{\sum_{n=1}^{k-1} v_{en}^2})
\end{align}
where $\eta$ is learning rate, $L_d$ equals to $3M$ and $L_e$ equals to $2N$. 
\end{theorem}

\paragraph{Derivation and Implementation}

We then demonstrate that the expected model update of \our{} above can be bounded with proper initialization. We provide the analysis on the encoder-only architecture, which can be naturally extended to encoder-decoder models in the same way. Analogous to \citet{HongyiZhang2019FixupIR} and \citet{deepnet}, we set our goal for the model update as follows:

\begin{framed}
\textbf{GOAL:} $F(x, \theta)$ is updated by $\Theta(\eta)$ per SGD step after initialization as $\eta \rightarrow 0$. That is $\Delta F^{sub} = \Theta(\eta d)$ where $\Delta F^{sub} \overset{\Delta}{=} F(x, \theta - \eta \frac{\partial \mathcal{L}}{\partial \theta}) - F(x, \theta)$.      
\end{framed}

Based on \cref{thm:deep_enc}, there are multiple methods to bound $\Delta F^{sub}$ independent of the depth by setting proper $v_l$ and $w_l$. In this work, we simply set $v_l = w_l = \gamma$ for all sub-layers. With \cref{eq:deep_enc_bound}, the term related to $L$ can be bounded as:

\begin{align}
    \cfrac{\sum_{l=1}^L(1 + \cfrac{v_l^2}{w_l^2})}{\sum_{n=1}^{L} v_n^2} 
    &+
    \cfrac{1}{\sum_{n=1}^{L} v_n^2} \sum_{l=1}^L\sum_{k=2}^L (1 + \cfrac{v_l^2}{w_l^2}) \cfrac{v_k^2}{\sum_{n=1}^{k-1} v_n^2}
    = \mathcal{O}(\cfrac{\log L}{\gamma^2}) \label{eq:impl}
\end{align}

We use $v = w = \gamma = \sqrt{\log L}$ to bound \cref{eq:impl} to $\mathcal{O}(1)$. In summary, we apply our initialization as follows:

\begin{tcolorbox}[enhanced,attach boxed title to top center={yshift=-3mm,yshifttext=-1mm}, title=\textbf{Encoder-only (or decoder-only) architecture}, colback=white, colframe=white!75!blue, coltitle=black, colbacktitle=white]
    \begin{enumerate}[leftmargin=*]
    \item Apply standard initialization (e.g., Xavier initialization) for each layer.
    \item For each layer, scale the weights of feed-forward networks as well as the value projection and the output projection of attention layers by $\sqrt{\log{2N}}$ (or $\sqrt{\log{2M}}$).
    \end{enumerate}
\end{tcolorbox}

The derivation of encoder-decoder architectures can be conducted in the same way (see \cref{appendix_derivation_en2de}). We summarize the steps as follows:

\begin{tcolorbox}[enhanced,attach boxed title to top center={yshift=-3mm,yshifttext=-1mm}, title=\textbf{Encoder-decoder architecture}, colback=white, colframe=white!75!blue, coltitle=black, colbacktitle=white]
    \begin{enumerate}[leftmargin=*]
    \item Apply standard initialization (e.g., Xavier initialization) for each encoder and decoder layer.
    \item For encoder layers, scale the weights of feed-forward networks as well as the value projection and the output projection of attention layers by $\sqrt{\frac{1}{3}\log{3M}\log{2N}}$.
    \item For decoder layers, scale the weights of feed-forward networks as well as the value projection and the output projection of attention layers by $\sqrt{\log{3M}}$.
    \end{enumerate}
\end{tcolorbox}

\section{Experiments on Language Tasks}

\subsection{Causal Language Modeling}
\label{gpt}

We implement \our{} on causal language modeling, which is the pretraining task for recent large language models (e.g., GPT-3~\citep{gpt3}, PaLM~\citep{palm}, etc). We start with a model that has the same model configuration as GPT-3 Medium (350M), and further scale its depth from 24L to 48L and 72L. The model is trained on an English-language corpus, which is a subset of the data from~\citet{roberta} and the English portion of CC100 corpus. We use the same tokenizer as GPT-2~\citep{gpt-2} to preprocess the data. The 24L model is trained for 500K steps, while the 48L and 72L models are trained for 250K steps. More details regarding the hyperparameters can be found in the appendix.

We compare \our{} with vanilla \preln{} Transformer and Normformer~\citep{Normformer2021}. Vanilla \preln{} is the backbone for GPT, while Normformer is a state-of-the-art model for causal language modeling. We use the implementation on the Fairseq\footnote{\url{https://github.com/facebookresearch/fairseq/}} codebase, and pre-train the models with the same monolingual data as described above.

We evaluate the performance of in-context learning. Following the previous work~\citep{gpt3,metalm}, we choose Winogrande~\citep{winogrande}, Winograd~\citep{wsc}, Storycloze~\citep{storycloze}, and Hellaswag~\citep{hellaswag} as the benchmark datasets, covering the cloze and completion tasks. We conduct experiments in the setting of zero-shot, one-shot, and four-shot learning. We randomly sample the examples from training data as demonstrations for the few-shot setting. The examples are concatenated with a separator \textit{</s>}.

\cref{tab:gpt_0_shot} summarizes the results in the zero-shot setting. It shows that \our{} achieves significant improvement over both vanilla \preln{} Transformer and Normformer. The improvement is consistent across different scales. Besides, it tolerates a larger learning rate than the baselines, indicating that \our{} is more stable in optimization. This allows the model to further scale up without pain. \cref{tab:gpt_1_shot} and \cref{tab:gpt_4_shot} report the results in the few-shot setting. \our{} is also better at few-shot learning than the baselines across four datasets, proving the effectiveness of \subln{} on causal language modeling.

\begin{table*}[t]
    \setlength{\tabcolsep}{11pt}
    \centering
    \begin{tabular}{l|c|c|cccc|c}
    \toprule
    \textbf{Models}  & \textbf{\# Layers} & \textbf{LR} & \textbf{WGe} & \textbf{WG} & \textbf{SC} & \textbf{HS} & \textbf{Avg.}\\
    \midrule
    \preln{} & \multirow{5}{*}{24L} & 5e-4 & \textbf{55.2} & 65.3 & 70.8 & 44.8 & 59.0  \\
    \preln{} & & 1e-3 & \multicolumn{5}{c}{diverged} \\
    Normformer & & 5e-4 & 54.3 & 68.1 & 72.0 & 45.9 & 60.1 \\
    Normformer & & 1e-3 & \multicolumn{5}{c}{diverged} \\
    \bf \our{} & & 1e-3 & 54.3 & \textbf{71.9} & \textbf{72.4} & \textbf{46.9} & \textbf{61.4} \\
    \midrule
    \preln{} & \multirow{3}{*}{48L} & 5e-4 & \textbf{57.3} & 67.0 & 74.0 & 48.0 & 61.6  \\
    Normformer & & 5e-4 & 56.5 & 70.5 & 74.0 & 49.8 & 62.7  \\
    \bf \our{} & & 1.2e-3 & 57.0 & \textbf{73.3} & \textbf{74.7} & \textbf{51.2} & \textbf{64.1} \\
    \midrule
    \preln{} & \multirow{3}{*}{72L} & 5e-4 & \textbf{58.0} & 70.9 & 75.7 & 51.7 & 64.1  \\
    Normformer & & 5e-4 & 57.4 & \textbf{75.4} & 75.2 & 53.6 & 65.4  \\
    \bf \our{} & & 1.2e-3 & 57.9 & 73.7 & \textbf{76.6}  & \textbf{55.1} & \textbf{65.8} \\
    \bottomrule
    \end{tabular}
    \caption{Zero-shot results for \our{} and the baselines (\texttt{WGe}: Winogrande, \texttt{WG}: Winograd, \texttt{SC}: Storycloze, and \texttt{HS}: Hellaswag dataset).}
    \label{tab:gpt_0_shot}
\end{table*}

\begin{table*}[t]
    \setlength{\tabcolsep}{11pt}
    \centering
    \begin{tabular}{l|c|c|cccc|c}
    \toprule
    \textbf{Models}  & \textbf{\# Layers} & \textbf{LR} & \textbf{WGe} & \textbf{WG} & \textbf{SC} & \textbf{HS} & \textbf{Avg.}\\
    \midrule
    \preln{} & \multirow{5}{*}{24L} & 5e-4 & \textbf{54.4} & 66.7 & 71.0 & 44.8 & 59.2 \\
    \preln{} & & 1e-3 & \multicolumn{5}{c}{diverged} \\
    Normformer & & 5e-4 & 54.0 & 67.4 & 72.1 & 45.6 & 59.8 \\
    Normformer & & 1e-3 & \multicolumn{5}{c}{diverged} \\
    \bf \our{} & & 1e-3 & 54.1 & \textbf{70.2} & \textbf{72.8} & \textbf{47.3} & \textbf{61.1} \\
    \midrule
    \preln{} & \multirow{3}{*}{48L} & 5e-4 & 56.0 & 69.5 & 74.2 & 48.5 & 62.1 \\
    Normformer & & 5e-4 & 54.7 & 71.2 & 74.8 & 50.6 & 62.8  \\
    \bf \our{} & & 1.2e-3 & \textbf{56.8} & \textbf{71.6} & \textbf{74.9} & \textbf{51.5} & \textbf{63.7} \\
    \midrule
    \preln{} & \multirow{3}{*}{72L} & 5e-4 & 56.9 & 71.2 & 76.0 & 52.2 & 64.1  \\
    Normformer & & 5e-4 & 57.8 & 69.8 & 76.8 & 54.0 & 64.6  \\
    \bf \our{} & & 1.2e-3 & \textbf{59.8} & \textbf{74.0} & \textbf{77.9} & \textbf{55.5} & \textbf{66.8} \\
    \bottomrule
    \end{tabular}
    \caption{One-shot results for \our{} and the baselines (\texttt{WGe}: Winogrande, \texttt{WG}: Winograd, \texttt{SC}: Storycloze, and \texttt{HS}: Hellaswag dataset).}
    \label{tab:gpt_1_shot}
\end{table*}

\begin{table*}[t]
    \setlength{\tabcolsep}{11pt}
    \centering
    \begin{tabular}{l|c|c|cccc|c}
    \toprule
    \textbf{Models}  & \textbf{\# Layers} & \textbf{LR} & \textbf{WGe} & \textbf{WG} & \textbf{SC} & \textbf{HS} & \textbf{Avg.}\\
    \midrule
    \preln{} & \multirow{5}{*}{24L} & 5e-4 & 54.0 & 67.7 & 69.8 & 44.6 & 59.0  \\
    \preln{} & & 1e-3 & \multicolumn{5}{c}{diverged} \\
    Normformer & & 5e-4 & 54.3 & 70.2 & 71.4 & 45.9 & 60.5 \\
    Normformer & & 1e-3 & \multicolumn{5}{c}{diverged} \\
    \bf \our{} & & 1e-3 & \textbf{57.6} & \textbf{74.7} & \textbf{72.8} & \textbf{47.5} & \textbf{63.2} \\
    \midrule
    \preln{} & \multirow{3}{*}{48L} & 5e-4 & 57.7 & 71.2 & 73.8 & 48.7 & 62.9  \\
    Normformer & & 5e-4 & 56.8 & \textbf{75.4} & 75.9 & 50.7 & \textbf{64.7}  \\
    \bf \our{} & & 1.2e-3 & \textbf{57.9} & 71.9 & \textbf{76.4} & \textbf{51.9} & 64.5 \\
    \midrule
    \preln{} & \multirow{3}{*}{72L} & 5e-4 & 57.5 & 73.3 & 76.1 & 52.4 & 64.8  \\
    Normformer & & 5e-4 & 57.7 & \textbf{74.0} & 77.0 & 54.9 & 65.9  \\
    \bf \our{} & & 1.2e-3 & \textbf{58.3} & \textbf{74.0} & \textbf{79.0} & \textbf{55.7} & \textbf{66.8} \\
    \bottomrule
    \end{tabular}
    \caption{Four-shot results for \our{} and the baselines (\texttt{WGe}: Winogrande, \texttt{WG}: Winograd, \texttt{SC}: Storycloze, and \texttt{HS}: Hellaswag dataset).}
    \label{tab:gpt_4_shot}
\end{table*}

\subsection{Masked Language Modeling}

We further conduct experiments on masked language modeling. We pre-train \our{} on a 16GB English corpus~\citep{roberta}, a combination of Wikipedia and Bookcorpus. We adopt the BERT-base setting and train a model with 12 layers, 768 hidden dimensions, and 3072 FFN dimensions. The batch size is 2048 and the model is trained for 125K steps. The vocabulary is built from a SentencePiece~\citep{sentencepiece} tokenizer with 64K tokens. More details are in the appendix.

We compare \our{} with both \postln{} and \preln{}. \postln{} is the de-facto standard for masked language modeling. We search the pre-training learning rate among \{5e-4, 1e-3, 2e-3, 3e-3\}, and choose the largest one that can converge. We fine-tune the models on the GLUE~\citep{glue} benchmarks. We run each experiment with three seeds and report the average results. \cref{tab:mlm} summarizes the results. It shows that \our{} has better performance than the strong baselines with a gain of average 0.6 points.

\begin{table*}[t]
\centering
\begin{tabular}{l|c|ccccccc|c}
\toprule
\textbf{Models} & \textbf{LR} & \textbf{MNLI} & \textbf{QNLI} & \textbf{QQP} & \textbf{SST} & \textbf{CoLA} & \textbf{MRPC} & \textbf{STS} & \textbf{Avg.} \\
\midrule
\postln{} & 5e-4 & \bf 86.7/86.7 & 92.2 & 91.0 & 93.4 & 59.8 & 86.4 & \bf 89.4 & 85.7 \\
\postln{} & 1e-3 & \multicolumn{8}{c}{diverged} \\
\preln{} & 1e-3 & 85.6/85.4 & 92.2 & 91.1 & 93.4 & 55.6 & 85.1 & 88.4 & 84.6 \\
\preln{} & 2e-3 & \multicolumn{8}{c}{diverged} \\
\midrule
\bf \our{}  & 3e-3 & \bf 86.7/86.7 & \bf 92.4 & \bf 91.2 & \bf 93.9 & \bf 62.9 & \bf 87.2 & 89.2 & \bf 86.3 \\
\bottomrule
\end{tabular}
\caption{Results on the GLUE development set.}
\label{tab:mlm}
\end{table*}

\subsection{Neural Machine Translation}
\label{nmt}

We also evaluate \our{} on machine translation. We perform experiments on OPUS-100 corpus, a multilingual machine translation dataset provided by~\citet{opus100}. OPUS-100 is an English-centric multilingual corpus covering 100 languages, which is randomly sampled from the OPUS collection. We implement \our{} with an 18-layer encoder, an 18-layer decoder, and 512 hidden dimension. We train the model with a batch size of 500K tokens for 100K steps. During testing, we select the checkpoint based on the performance of the validation set. We use the beam search algorithm with a beam size of 5 and set the length penalty as 1.0.  More details are in the
appendix.

\cref{tab:mt} reports the BLEU scores on the OPUS-100 test sets. \postln{} can not converge with the depth of 18L-18L due to the training instability. \preln{} is the standard alternative when the model is deep and large. Compared to \preln{} and its variant Normformer, \our{} has an improvement of average 0.5 and 0.6 BLEU scores, proving the effectiveness on the machine translation task.

\begin{table*}[t]
\begin{center}
\begin{tabular}{l|cc|c}
\toprule
\textbf{Models} & \textbf{En $\rightarrow$ X} & \textbf{X $\rightarrow$ En} & \textbf{Avg.}  \\
\midrule
\postln{} & \multicolumn{3}{c}{diverged} \\
\preln{} & 28.3	& 32.7	&  30.5 \\
NormFormer & 28.5 & 32.3 & 30.4
\\
\midrule 
 \bf \our{} & \textbf{28.7} & \textbf{33.2} & \textbf{31.0} 
 \\
\bottomrule
\end{tabular}
\caption{BLEU scores for \our{} and the baselines on the OPUS-100 test sets.}
\label{tab:mt}
\end{center}
\end{table*}

\section{Experiments on Vision Tasks}

\begin{table*}[t]
\centering
\begin{tabular}{l|c|cccc|c}
\toprule
\bf \multirow{2}{*}{Models}  & \bf \multirow{2}{*}{\# Layers} & \bf \multirow{2}{*}{ImageNet} & \bf ImageNet & \bf ImageNet & \bf ImageNet & \bf \multirow{2}{*}{ADE20k} \\
& & & \bf Adversarial & \bf Rendition & \bf Sketch & \\
\midrule
\preln{} & \multirow{2}{*}{12L}  &  84.5 & 45.9 & 55.6 & 42.2 & 51.4  \\
\bf \our{} &  & \bf 84.9 & \bf 48.9 & \bf 57.7 & \bf 43.9 & \bf 52.2 \\
\midrule
\preln{} & \multirow{2}{*}{24L} & 86.2 & 60.1 & 63.2 & 48.5 & 54.2 \\
\bf \our{} &  & \bf 86.8 & \bf 65.4 & \bf 67.5 & \bf 52.0 & \bf 54.6 \\
\bottomrule
\end{tabular}
\caption{Results on vision tasks. \preln{} is instantiated as vanilla ViT~\citep{vit}. We report top-1 accuracy on ImageNet and its variants, and mIoU metric on ADE20k for semantic segmentation. We compare both ViT-Base (12L) and ViT-Large (24L).}
    \label{tab:vision_task}
\end{table*}

We pretrain \our{} under masked image modeling framework (BEiT; \citealt{beit}; \citealt{beit2}), and then fine-tune it on various downstream vision tasks by appending lightweight task layers.
To be specific, we encourage \our{} to reconstruct corresponding discrete visual tokens~\citep{beit2}, based on the corrupt input images. 

In comparison, \preln{} is instantiated as vanilla ViT~\citep{vit} here and pretrained under the same settings.
We pretrain all models on ImageNet-1k~\citep{imagenet} with 300 epochs schedule. 
After that, we fine-tune the pretrained models on ImageNet-1k for the image classification task and on ADE20k~\citep{ade20k} for the semantic segmentation task.
Moreover, we evaluate the robustness of all fine-tuned models on various ImageNet variants, \eg{}, ImageNet-Adversarial~\citep{adversarial2021}, ImageNet-Rendition~\citep{rendition2021} and
ImageNet-Sketch~\citep{sketch2019}.
We summarize the results of those vision tasks in Table~\ref{tab:vision_task}.
Hyperparameters are given in Appendix~\ref{app:hyperparam}.

As shown in Table~\ref{tab:vision_task}, \our{} outperforms its \preln{} counterpart by 0.4\% and 0.6\% when the number of layers is 12 and 24 on ImageNet validation set, respectively.
Moreover, \our{} outperforms ViT by a significant margin across three ImageNet variants.
By appending the UperNet~\citep{upernet} task layer, we conduct semantic segmentation experiments on ADE20k.
For 12-layer models, \our{} reach 52.2\% mIoU, which is 0.8\% higher than vanilla ViT.
For 24-layer models, \our{} can boost the performance to 54.6\%.

\section{Experiments on Speech Tasks}

We implement the proposed \our{} based on the open-source ESPnet repository \citep{watanabe2018espnet} for speech recognition, and evaluate its performance on the LibriSpeech 960h \citep{panayotov2015librispeech}  benchmark.  

Since the transducer framework is proven to obtain better accuracy with low latency, we choose the Transformer Transducer (T-T; \citealt{zhang2020transformer}) as the backbone framework, where the encoder is either Pre-LN Transformer or  \our{}, and the predictor network is a two-layer LSTM network. The model input is 80 dimension filter bank feature and its output vocabulary is 5000 subword units. There is a VGG component before Transformer blocks to downsample the speech frame rate from 10 to 40 milliseconds. 

We evaluate 18L and 36L T-T with  hidden state dimensions of 512 and FFN dimensions of 2048. Their numbers of parameters are 80M and 140M respectively.  The models are trained for 150 epochs on the full 960 hours of audio data in LibriSpeech, where the adaptive specaugement \citep{specaugment,DBLP:conf/icassp/ParkZCCLCLW20} is employed for data augmentation. The auxiliary loss proposed in \cite{boyer2021study} is used for better performance. 
Table \ref{tab:libri_asr} shows the evaluation results on \texttt{dev-clean},  \texttt{dev-other}, \texttt{test-clean}, and \texttt{test-other}. \our{} achieves over 6$\%$ WER reduction against the Transformer baseline in the 18L setting. A similar gain is also observed in the 36L setting. When searching for the best learning rate, we find that 36L \our{} allows a learning rate up to 3e-3, while Transformer can only be trained with $lr=1.5e-3$. Regarding the 18L setting, \our{} and Pre-LN are trained with $lr=5e-3$ and $lr=3e-3$, respectively.

 \begin{table*}[t] 
    \centering
    \begin{tabular}{l|c|cccc}
    \toprule
    \bf {Models}  & \bf {\# Layers} & \bf {Dev-Clean} & \bf Dev-Other & \bf Test-Clean & \bf Test-Other  \\
    \midrule
    \preln{} & \multirow{2}{*}{18L}  & 2.97	&6.52	&3.19	&6.62  \\
    \bf \our{} &  & \bf 2.68	&\bf 6.04&	\bf 2.99	&\bf 6.16   \\
    \midrule
    \preln{} & \multirow{2}{*}{36L} & 2.59	&6.10&	2.89	&6.04  \\
    \bf \our{} &  & \bf 2.43 &\bf 	5.34	&\bf 2.72	&\bf 5.56 \\
    \bottomrule
    \end{tabular}
    \caption{Results on speech recognition. All models are without language model shallow fusion.}
    \label{tab:libri_asr}
\end{table*}

\section{Experiments on Vision-Language Tasks}

We conduct experiments on multimodal pretraining following BEiT-3~\citep{beit3} and evaluate the model on downstream vision-language benchmarks, including VQA 2.0~\citep{vqa} and NLVR2~\citep{nlvr2}.
Specifically, we perform masked data modeling on images, texts and image-text pairs to learn multimodal representations.
We compare \our{} with the \preln{} variant as in vanilla ViT~\citep{vit} under the same pretraining setting.
We pretrain a 24-layer base model with 544 hidden dimensions and 2176 FFN dimensions using the same pretraining data as in BEiT-3.
The peak learning rate is 2e-3 and the batch size is 12,288 for \our{} and the baseline.
Each batch contains 4096 images, 4096 texts and 4096 image-text pairs.
Both two models are trained for 300k steps.

As present in Table~\ref{tab:vision_language_task}, \our{} achieves consistent improvements across two vision-language benchmarks.
\our{} outperforms standard \preln{} by 0.5\% on VQA test-standard split and NLVR2 test set.

\begin{table*}[t]
\centering
\begin{tabular}{l|c|cc|cc}
\toprule
\bf \multirow{2}{*}{Models} & \bf \multirow{2}{*}{\# Layers} & \multicolumn{2}{c|}{\bf VQA} & \multicolumn{2}{c}{\bf NLVR2} \\
& & test-dev & test-std & dev & test-P \\
\midrule
\preln{} & \multirow{2}{*}{24L} & 78.37 & 78.50 & 82.57 & 83.69 \\
\bf \our{} &  & \bf 79.00 & \bf 79.01 & \bf 83.35 & \bf 84.23 \\
\bottomrule
\end{tabular}
\caption{Results on vision-language tasks. 
We report vqa-score on VQA test-dev and test-standard split, as well as accuracy on NLVR2 development and public test set (test-P).
}
\label{tab:vision_language_task}
\end{table*}

\section{Conclusion}

In this paper, we call for the development of Foundation Transformers, and present \our{}, an implementation of Foundation Transformers towards a true general-purpose architecture across various tasks and modalities. Experiments demonstrate that \our{} achieves better results than the baselines on language, vision, speech, and multimodal tasks. More importantly, \our{} has theoretically-guaranteed training stability which makes it a promising option for scaling up any Transformer models.

\bibliographystyle{plainnat}
\bibliography{deepnet}

\newpage

\appendix

\section{Model update for Encoder-only Transformers}

\subsection{\preln{}}
\label{sec:pre}

Following~\citet{deepnet}, query and key projection do not impact the bound of model update's magnitude. We thus only consider the re-scaling effect of input and output projection in feed-forward layers, value and output projection in attention layers. The forward propagation for an $N$-layer \preln{} Transformer based on encoder-only architecture is:

\begin{align}
    &F(x; \theta) = W^{vocab} x^{e} \\
    &x^{e} = \text{LN}(x + \sum_{l=1}^{L} G^l(x^{l-1}, \theta_{el})), 
    \quad x^l = G^l(x^{l-1}, \theta_{el}) \\
    &x^0 = x,\,x_i \backsim \mathcal{N}(0, 1)\,\text{and}\,W_{ij}^{vocab} \backsim \mathcal{N}(0, \cfrac{1}{d})
\end{align}

$\theta_e$ denotes the parameters of output projection $W^{vocab}$ and backbone $\{\theta_{el}\}_{l=1}^L$. $W^o \in R^{V \times d}$, where $d$ is hidden dimension. $L$ equals to $2N$ for simplicity. Without the loss of generality, we set the intermediate dimension of feed-forward layers equals to hidden dimension. The forward computation of $l$-th sub-layer can be formulated as follows:

\begin{equation}
\label{eq:pre_for1}
    x_i^{l} = \sum_{j=1}^d W_{ij}^{l,2} u_j^{l} + x_i^{l-1}
\end{equation}

\begin{equation}
\label{eq:pre_for2}
    u_i^l = \phi(z_i^l)
\end{equation}

\begin{equation}
\label{eq:pre_for3}
    z_i^{l} = \sum_{j=1}^d W_{ij}^{l,1} \text{LN}_j(x^{l-1}) = \sum_{j=1} W_{ij}^{l,1}\cfrac{x_j^{l-1} - \cfrac{1}{d}\sum_{k=1}^d x_k^{l-1}}{\sqrt{\cfrac{1}{d}\sum_{k=1}^d (x_k^{l-1} - \overset{-}{x^{l-1}})^2}}
\end{equation}

$x_i^{l-1}$ and $x_i^{l}$ is $i$-th entry of input and output vector respectively. $\phi$ refers to activation function. $W_{ij}^{l,1}$, $W_{ij}^{l,2}$ denotes the $i$-th row, $j$-th column entry of input and output projection for feed-forward layer, or value and output projection for attention layer. We first perform Xavier initialization for all parameters, then re-scale them with a constant. For example, $W_{ij}^{l,1}$, $W_{ij}^{l,2}$ satisfies that:

\begin{equation}
    W_{ij}^{l,1} \backsim \mathcal{N}(0, \frac{w_l^2}{d}), \quad W_{ij}^{l,2} \backsim \mathcal{N}(0, \frac{v_l^2}{d})
\end{equation}

$v_l$ and $w_l$ are factors for re-scaling after standard initialization. For vanilla \preln{} Transformer, $v_l$ and $w_l$ equal to 1.

By means of Taylor expansion, we ignore the second-order term. Model update $\Delta F$ satisfies that:

\begin{equation}
    \Delta F = \sum_{i=1}^d \cfrac{\partial F}{\partial x_i^e} \cfrac{\partial x_i^e}{\partial W}
\end{equation}

To simplify the derivation, we make following assumption: for $i$-th entry of backbone output $x^e$, we only consider the update of corresponding entry of each sub-layer's output $x^l$, which means that $\frac{\partial x_i^e}{\partial x_j^l}$ equals to 0 when $i \ne j$.

With \cref{eq:pre_for1}, \cref{eq:pre_for2} and \cref{eq:pre_for3}, we estimate the magnitude of $\frac{\partial x_i^e}{\partial W_{ij}^{l,2}}$ and $\frac{\partial x_i^e}{\partial W_{ij}^{l,1}}$. For simplicity, we omit the index of output, i.e., $x_i^e = x^e$ in the following.

\begin{align}
\label{eq:grad_pre_w2}
    \cfrac{\partial x^e}{\partial W_{ij}^{l,2}} = \delta_i^{l} u_j^{l},\quad \delta_i^{l} = \cfrac{\partial x^e}{\partial G_{i}^{l}}
\end{align}

\begin{align}
    \cfrac{\partial x^e}{\partial W_{mn}^{l,1}} 
    = \cfrac{\partial x^e}{\partial G_i^l} \cfrac{\partial G_i^l}{\partial u_m^l}\cfrac{\partial u_m^l}{\partial z_m^l} LN_n(x^{l-1})
    \overset{\Theta}{=} \delta_i^l W_{im}^{l,2}
\end{align}

Since the magnitude of the gradients which goes through more than two layer normalization converges as the depth $L$ grows, for $\delta_k^l$ we consider the magnitude of $\frac{\partial x^e}{\partial G_i^l}$ and $\sum_{k=l+1}^L\frac{\partial x^e}{\partial G_i^k} \frac{\partial G_i^k}{\partial G_i^l}$. With $\frac{\partial \text{LN}(x)}{\partial x} = \mathcal{O}(\frac{\sqrt{d}}{||x||_2})$, the magnitude of $\delta_k^l$ satisfies that:

\begin{align}
\label{eq:delta_l}
    &\delta_k^l
    \overset{\Theta}{=}
    (1 + \sum_{k=l+1}^L \cfrac{v_k w_k}{\sqrt{\sum_{n=1}^{k-1} v_n^2 w_n^2}})\cfrac{1}{\sqrt{\sum_{n=1}^{L} v_n^2 w_n^2}} = \delta^l, \quad 1\le l \le L-1\\
    &\delta_k^L \overset{\Theta}{=} \cfrac{1}{\sqrt{\sum_{n=1}^{L} v_n^2 w_n^2}}
\end{align}

We have the bounds of model update caused by $W^2 = \{ W^{l,2} \}_{l=1}^L$ and $W^1 = \{ W^{l,1} \}_{l=1}^L$:

\begin{align}
    &\Delta F_{W^2} 
    = \sum_{l=1}^L \sum_{i,j}^d \cfrac{\partial F}{\partial x_i^e} \cfrac{\partial x_i^e}{\partial W_{ij}^{l,2}} \Delta W_{ij}^{l,2} 
    = \sum_{l=1}^L \sum_{i,j}^d \delta^l u_j^l W_i^{vocab} \Delta W_{ij}^{l,2} \\
    &\Delta F_{W^1} 
    = \sum_{l=1}^L \sum_{i,m,n}^d \cfrac{\partial F}{\partial x_i^e}\cfrac{\partial x_i^e}{\partial W_{mn}^{l,1}} \Delta W_{mn}^{l,1}
    = \sum_{l=1}^L \sum_{i,m,n}^d \delta^l W_{im}^{l,2} W_i^{vocab} \Delta W_{mn}^{l,1}\\
\end{align}

Then we estimate $\Delta F$ under SGD update. Following \citet{fim}, we introduce $\overset{-}{p}^l$ and $\overset{-}{q}^l$ for forward and backward signal propagation of $l$-th sub-layer.

\begin{align}
    &\overset{-}{q}^l 
    = \sum_{i=1}^d (\delta_i^l)^2
    \overset{\Theta}{=} \cfrac{d}{\sum_{n=1}^{L} v_n^2 w_n^2}(1 + \sum_{k=l+1}^L \cfrac{v_k^2 w_k^2}{\sum_{n=1}^{k-1} v_n^2 w_n^2}) \label{eq:q_l} \\
    &\overset{-}{p}^l = \cfrac{1}{d}\sum_{j=1}^d (u_j^l)^2 \overset{\Theta}{=} w_l^2
\end{align}

Above all, we have the bound for $N$-layer \preln{} Transformer's update $\Delta F$, where $\eta$ is learning rate:

\begin{align}
    \Delta F &= \Delta F_{W^1} + \Delta F_{W^2} = \eta \sum_{l=1}^L (v_l^2 + w_l^2) \overset{-}{q}^l \\
    &\overset{\Theta}{=} \eta d(\cfrac{\sum_{l=1}^L v_l^2 + w_l^2}{\sum_{n=1}^{L} v_n^2 w_n^2} + \sum_{l=1}^L \sum_{k=2}^L\cfrac{v_l^2 + w_l^2}{\sum_{n=1}^{L} v_n^2 w_n^2}\,\cfrac{v_k^2 w_k^2}{\sum_{n=1}^{k-1} v_n^2 w_n^2}))
\end{align}

\subsection{\our{}}

We give theoretical analysis in the following section. For an $N$-layer, encoder-only \our{}, the forward computation of the $l$-th sub-layer can be formulated as:

\begin{equation}
\label{eq:dep_pre1}
    x_i^{l} = \sum_{j=1}^d W_{ij}^{l,2} u_j^{l} + x_i^{l-1}
\end{equation}

\begin{equation}
\label{eq:dep_pre2}
    u_i^l = \text{LN}(\phi(z_i^l))
\end{equation}

\begin{equation}
\label{eq:dep_pre3}
    z_i^{l} = \sum_{j=1}^d W_{ij}^{l,1} \text{LN}_j(x^{l-1})
\end{equation}

Following the same assumptions in \cref{sec:pre}, the gradient $\frac{\partial x^e}{\partial W_{ij}^{l,2}}$ is the same as it in \cref{eq:grad_pre_w2}. With \cref{eq:dep_pre1}, \cref{eq:dep_pre2} and \cref{eq:dep_pre3}, we estimate $\frac{\partial x^e}{\partial W_{mn}^{l,1}}$ as follows:

\begin{align}
    \cfrac{\partial x^e}{\partial W_{mn}^{l,1}} 
    = \cfrac{\partial x^e}{\partial G_i^l} \cfrac{\partial G_i^l}{\partial u_m^l}\cfrac{\partial u_m^l}{\partial z_m^l} \text{LN}_n(x^{l-1})
    \overset{\Theta}{=} \cfrac{\delta_k^l}{w_l} W_{ki}^{l,2}
\end{align}

It is noted that with additional normalization, re-scaling factor $w_l$ of input projection does not impact the magnitude of sublayer's output $G^l$, and $\overset{-}{p}^l$ is normalized to 1. Therefore, we have the bound of the magnitude of $\delta_k^l$ and $\overset{-}{q}^l$:

\begin{align}
    &\delta_k^l
    \overset{\Theta}{=}
    (1 + \sum_{k=l+1}^L \cfrac{v_k }{\sqrt{\sum_{n=1}^{k-1} v_n^2}}) \cfrac{1}{\sqrt{\sum_{n=1}^{L} v_n^2}}, \quad 1 \le l \le L-1\\
    &\delta_k^L = \cfrac{1}{\sqrt{\sum_{n=1}^{L} v_n^2}} \\
    &\overset{-}{q}^l
    \overset{\Theta}{=} \cfrac{d}{\sum_{n=1}^{L} v_n^2}(1 + \sum_{k=l+1}^L \cfrac{v_k^2}{\sum_{n=1}^{k-1} v_n^2})
\end{align}

We have the bound of model update caused by $W^1$ and $W^2$ under SGD respectively:

\begin{align}
    \Delta F_{W^2} = \eta \sum_{l=1}^L \overset{-}{q}^l, \quad 
    \Delta F_{W^1} = \eta \sum_{l=1}^L\cfrac{v_l^2}{w_l^2}\overset{-}{q}^l
\end{align}

Above all, the bound of the model update's magnitude $\Delta F$ satisfies that:

\begin{align}
    \Delta F &= \Delta F_{W^1} + \Delta F_{W^2} 
    = \eta \sum_{l=1}^L (1 + \cfrac{v_l^2}{w_l^2}) \overset{-}{q}^l \\
    &\overset{\Theta}{=} \eta d(\cfrac{\sum_{l=1}^L 1 + \cfrac{v_l^2}{w_l^2}}{\sum_{n=1}^{L} v_n^2} +
    \cfrac{1}{\sum_{n=1}^{L} v_n^2} \sum_{l=1}^L\sum_{k=2}^L (1 + \cfrac{v_l^2}{w_l^2}) \cfrac{v_k^2}{\sum_{n=1}^{k-1} v_n^2})
\end{align}

\section{Model update for Encoder-decoder Transformers}

\subsection{\preln{}}

The derivation of self-attention and FFN layers is given in \cref{sec:pre}. For $l$-th cross attention layer, the forward computation is:

\begin{equation}
    y_i^{l} = \sum_{j=1}^d W_{ij}^{l,2} u_j^{l} + y_i^{l-1}
\end{equation}

\begin{equation}
    u_i^l = \phi(z_i^l)
\end{equation}

\begin{equation}
    z_i^{l} = \sum_{j=1}^d W_{ij}^{l,1} x_j^e
\end{equation}

$x^e$ is the output of the encoder. $\delta_d^l$ and $\overset{-}{q}_d^l$ are given in \cref{eq:delta_l} and \cref{eq:q_l} respectively. Then we estimate the bound of $\frac{\partial f}{\partial x_j^e}$:

\begin{align}
    &\cfrac{\partial F}{\partial x_j^e} 
    \overset{\Theta}{=} \sum_{l=1, l\%3=1}^{L_d}\cfrac{\partial F}{\partial y_i^d}\cfrac{\partial y_i^d}{\partial y_i^l}\cfrac{\partial y_i^l}{\partial x_j^e} 
    \overset{\Theta}{=}
    \sum_{l=1, l\%3=1}^{L_d} W_i^{vocab}\delta_i^{l} \sum_{k=1}^d W_{ik}^{l,2} \sum_{j=1}^d W_{kj}^{l,1}
\end{align}

The bound of $||\cfrac{\partial F}{\partial x^e}||_2^2$ satisfies that:

\begin{align}
    &||\cfrac{\partial F}{\partial x^e}||_2^2 
    = \sum_{j=1}^d (\cfrac{\partial F}{\partial x_j^e})^2
    \overset{\Theta}{=} \sum_{l=1, l\%3=1}^{L_d} \cfrac{v_l^2 w_l^2}{d} \overset{-}{q}_d^l
\end{align}

Above all, under SGD update, we have the model update $\Delta F_{ed}$ for a $N$-layer encoder, $M$-layer decoder \preln{} Transformer:

\begin{align}
    &\Delta F_{ed} \le \Delta F_{d} + \sum_{l=1, l\%3=1}^{L_d} \cfrac{v_{dl}^2 w_{dl}^2}{\sum_{n=1}^{L_d} v_{dn}^2 w_{dn}^2}(1 + \sum_{k=2}^{L_d} \cfrac{v_{dk}^2 w_{dk}^2}{\sum_{n=1}^{k-1} v_{dn}^2 w_{dn}^2}) \Delta F_{e} \\
    &\Delta F_{d} 
    \overset{\Theta}{=} \eta d(\cfrac{\sum_{l=1}^{L_d} v_{dl}^2 + w_{dl}^2}{\sum_{n=1}^{L_d} v_{dn}^2 w_{dn}^2} + \sum_{l=1}^{L_d} \sum_{k=2}^{L_d}\cfrac{v_{dl}^2 + w_{dl}^2}{\sum_{n=1}^{L_d} v_{dn}^2 w_{dn}^2}\,\cfrac{v_{dk}^2 w_{dk}^2}{\sum_{n=1}^{k-1} v_{dn}^2 w_{dn}^2})) \\
    &\Delta F_{e} 
    \overset{\Theta}{=} \eta d(\cfrac{\sum_{l=1}^{L_e} v_{el}^2 + w_{el}^2}{\sum_{n=1}^{L_e} v_{en}^2 w_{en}^2} + \sum_{l=1}^{L_e} \sum_{k=2}^{L_e}\cfrac{v_{el}^2 + w_{el}^2}{\sum_{n=1}^{L_e} v_{en}^2 w_{en}^2}\,\cfrac{v_{ek}^2 w_{ek}^2}{\sum_{n=1}^{k-1} v_{en}^2 w_{en}^2}))
\end{align}

where $L_d$ equals to $3M$, $L_e$ equals to $2N$.

\subsection{\our{}}
\label{appendix_derivation_en2de}

The forward computation of cross attention layer for \our{} is:

\begin{equation}
    y_i^{l} = \sum_{j=1}^d W_{ij}^{l,2} u_j^{l} + y_i^{l-1}
\end{equation}

\begin{equation}
    u_i^l = \text{LN}(\phi(z_i^l))
\end{equation}

\begin{equation}
    z_i^{l} = \sum_{j=1}^d W_{ij}^{l,1} x_j^e
\end{equation}

Similarly we estimate the bound of $||\cfrac{\partial F}{\partial x^e}||_2^2$:

\begin{align}
    &\cfrac{\partial F}{\partial x_j^e} \overset{\Theta}{=} 
    \sum_{l=1, l\%3=1}^{L_d}\cfrac{\partial F}{\partial y_i^l}\cfrac{\partial y_i^l}{\partial x_j^e} \overset{\Theta}{=}
    \sum_{l=1, l\%3=1}^{L_d} W_i^{vocab} \delta_i^{l} \sum_{k=1}^d W_{ik}^{l,2} \sum_{j=1}^d \cfrac{\sqrt{d}}{||\phi(z^l)||} W_{kj}^{l, 1}\\
    &||\cfrac{\partial F}{\partial x^e}||_2^2
    = \sum_{j=1}^d (\cfrac{\partial F}{\partial x_j^e})^2
    \overset{\Theta}{=} \sum_{l=1, l\%3=1}^{L_d} \cfrac{v_l^2}{d} \overset{-}{q}_d^l \label{eq:deep_pre_ed}
\end{align}

With \cref{eq:deep_pre_ed}, we have the bound of the model update $\Delta F_{ed}$ for a $N$-layer encoder, $M$-layer decoder \our{}:

\begin{align}
    &\Delta F_{ed} \le \Delta F_{d} + \sum_{l=1, l\%3=1}^{L_d} \cfrac{v_{dl}^2}{\sum_{n=1}^{L_d} v_{dn}^2}(1 + \sum_{k=2}^{L_d} \cfrac{v_{dk}^2}{\sum_{n=1}^{k-1} v_{dn}^2}) \Delta F_{e} \label{eq:bound_en2de} \\
    &\Delta F_{d} \overset{\Theta}{=} \eta d(\cfrac{\sum_{l=1}^{L_d}(1 +  \cfrac{v_{dl}^2}{w_{dl}^2})}{\sum_{n=1}^{L_d} v_{dn}^2} +
    \cfrac{1}{\sum_{n=1}^{L_d} v_{dn}^2} \sum_{l=1}^{L_d}\sum_{k=2}^{L_d} (1 + \cfrac{v_{dl}^2}{w_{dl}^2}) \cfrac{v_{dk}^2}{\sum_{n=1}^{k-1} v_{dn}^2}) \\
    &\Delta F_{e} \overset{\Theta}{=} \eta d(\cfrac{\sum_{l=1}^{L_e}(1 + \cfrac{v_{el}^2}{w_{el}^2})}{\sum_{n=1}^{L_e} v_{en}^2} +
    \cfrac{1}{\sum_{n=1}^{L_e} v_{en}^2} \sum_{l=1}^{L_e}\sum_{k=2}^{L_e} (1 + \cfrac{v_{el}^2}{w_{el}^2}) \cfrac{v_{ek}^2}{\sum_{n=1}^{k-1} v_{en}^2})
\end{align}

There are multiple methods to bound $\Delta F_{ed}$ independent of the depth by setting proper $v_{el}$, $w_{el}$, $v_{dl}$ and $w_{dl}$. In this work, we set $v_{el} = w_{el} = \gamma_e$ and $v_{dl} = w_{dl} = \gamma_d$ for all sub-layers. We first use $\gamma_d = \sqrt{\log 3M}$ to bound $\Delta F_{d}$ to $\mathcal{O}(\eta d)$. With $\gamma_d = \sqrt{\log 3M}$, the second term of \cref{eq:bound_en2de} satisfies that:

\begin{align}
    \sum_{l=1, l\%3=1}^{L_d} \cfrac{v_{dl}^2}{\sum_{n=1}^{L_d} v_{dn}^2}(1 + \sum_{k=2}^{L_d} \cfrac{v_{dk}^2}{\sum_{n=1}^{k-1} v_{dn}^2}) \Delta F_{e} = \mathcal{O}(\cfrac{\log 3M \log 2N}{3\gamma_e^2}) = \mathcal{O}(1)
\end{align}

It leads to $\gamma_e = \sqrt{\frac{1}{3} \log 3M \log 2N}$.

\newpage
\section{Hyperparameters}
\label{app:hyperparam}

\begin{table}[ht]
\centering
\small
\scalebox{0.98}{
\begin{tabular}{l|ccc}
\toprule
\bf Hyperparameters & \bf Base Size & \bf Large Size & \bf Xd Size\\
\midrule
Layers & 24 & 48 & 72 \\
Hidden size & \multicolumn{3}{c}{1024} \\
FFN inner hidden size & \multicolumn{3}{c}{3072} \\
Attention heads & \multicolumn{3}{c}{16} \\
\midrule
Training updates & 500K & \multicolumn{2}{c}{250K} \\
Peak learning rate & \multicolumn{3}{c}{ \{5e-4, 7e-4, 1e-3, 1.2e-3\}} \\
Tokens per sample & \multicolumn{3}{c}{2048} \\
Batch size & \multicolumn{3}{c}{256} \\
Adam $\beta$ & \multicolumn{3}{c}{(0.9, 0.98)} \\
Learning rate schedule & \multicolumn{3}{c}{Polynomial decay} \\
Warmup updates & \multicolumn{3}{c}{750} \\
\midrule
Gradient clipping & \multicolumn{3}{c}{\xmark} \\
Dropout & \xmark & \multicolumn{2}{c}{0.1} \\
Attention dropout & \xmark & \multicolumn{2}{c}{0.1} \\
Weight decay & \multicolumn{3}{c}{0.01} \\
\bottomrule
\end{tabular}
}
\caption{
Hyperparameters for \our{} and the baselines pre-training on causal language modeling. 
}
\label{tbl:hyper:lang:gpt}
\end{table}

\begin{table}[ht]
\centering
\small
\scalebox{0.98}{
\begin{tabular}{l|c}
\toprule
\bf Hyperparameters & \bf MLM pretraining \\
\midrule
Layers & 12 \\
Hidden size & 768 \\
FFN inner hidden size & 3072 \\
Attention heads & 12 \\
\midrule
Peak Learning rate & \{5e-4, 1e-3, 2e-3, 3e-3\} \\
Learning rate schedule & Polynomial decay \\
Warm-up updates & 10,000 \\
Warm-up init learning rate & 1e-7 \\
Tokens per sample & 512 \\
Batch size & 2048 \\
Mask ratio & 15\% \\
Adam $\beta$ & (0.9, 0.98) \\
Training updates & 125K \\
\midrule
Gradient clipping & 2.0 \\
Dropout & 0.1 \\
Weight decay & \xmark \\
\bottomrule
\end{tabular}
}
\caption{
Hyperparameters for \our{} and the baselines on masked language model pretraining. 
}
\label{tbl:hyper:lang:pt}
\end{table}

\begin{table}[ht]
\centering
\small
\scalebox{0.98}{
\begin{tabular}{l|cc}
\toprule
\bf Hyperparameters & \bf Large Task & \bf Small Task \\
\midrule
Peak Learning rate & \multicolumn{2}{c}{\{1e-5, 2e-5, 3e-5, 4e-5, 1e-4, 2e-4, 3e-4, 4e-4\}} \\
Adam $\beta$ & \multicolumn{2}{c}{(0.9, 0.98)} \\
Warm-up & \{10\%, 20\%\} & \{10\%, 16\%\} \\
Batch size & 32 &  \{16, 32\} \\
Training epochs & 3 & \{2, 3, 5, 10\} \\
Seed & \multicolumn{2}{c}{\{1, 2, 3\}} \\
\midrule
Gradient clipping & \multicolumn{2}{c}{\xmark} \\
Dropout & \multicolumn{2}{c}{0.1} \\
Weight decay & \multicolumn{2}{c}{0.01} \\
\bottomrule
\end{tabular}
}
\caption{
Hyperparameters for \our{} and the baselines fine-tuning on the GLUE benchmark. (Large tasks include MNLI, QNLI, QQP, and SST. Small tasks are CoLA, MRPC, and STS.)
}
\end{table}

\begin{table}[ht]
\centering
\small
\scalebox{0.98}{
\begin{tabular}{l|c}
\toprule
\bf Hyperparameters & \bf Base Size \\
\midrule
Layers & 18L-18L \\
Hidden size & 512 \\
FFN inner hidden size & 2048 \\
Attention heads & 8 \\
\midrule
Peak Learning rate & 4e-3 \\
Learning rate schedule & Inverse sqrt \\
Warm-up updates & 8,000 \\
Warm-up init learning rate & 1e-7 \\
Max tokens & 128 $\times$ 4K \\
Adam $\beta$ & (0.9, 0.98) \\
Label smoothing & 0.1 \\
Training updates & 100K \\
\midrule
Gradient clipping & 1.0 \\
Dropout & 0.1 \\
Weight decay & \xmark \\
\bottomrule
\end{tabular}
}
\caption{
Hyperparameters for \our{} and the baselines on the machine translation. 
}
\label{tbl:hyper:lang:mt}
\end{table}

\begin{table}[ht]
\centering
\small
\scalebox{0.98}{
\begin{tabular}{l|cc}
\toprule
\bf Hyperparameters &  \multicolumn{2}{c}{\bf BEiT pretraining} \\
\midrule
Layers & 12 & 24 \\
Hidden size & 768 & 1024 \\
FFN inner hidden size & 3072 & 4096 \\
Attention heads & 12 & 16 \\
Patch size & \multicolumn{2}{c}{$16 \times 16$} \\
\midrule
Training epochs & \multicolumn{2}{c}{300} \\
Batch size & \multicolumn{2}{c}{2048} \\
Adam $\beta$ & \multicolumn{2}{c}{(0.9, 0.98)} \\
Peak learning rate & \multicolumn{2}{c}{1.5e-3} \\
Minimal learning rate & \multicolumn{2}{c}{1e-5} \\
Learning rate schedule & \multicolumn{2}{c}{Cosine} \\
Warmup epochs & \multicolumn{2}{c}{10} \\
\midrule
Gradient clipping & \multicolumn{2}{c}{3.0} \\
Dropout & \multicolumn{2}{c}{\xmark} \\
Drop path & \multicolumn{2}{c}{0} \\
Weight decay & \multicolumn{2}{c}{0.05} \\
\midrule
Data Augment & \multicolumn{2}{c}{RandomResizeAndCrop} \\
Input resolution & \multicolumn{2}{c}{$224 \times 224$} \\
Color jitter & \multicolumn{2}{c}{0.4} \\
\bottomrule
\end{tabular}
}
\caption{
Hyperparameters for \our{} pretraining on ImageNet-1K. 
}
\label{tbl:hyper:vision:pretrain}
\end{table}

\begin{table}[ht]
\centering
\scalebox{0.95}{
\begin{tabular}{l|cc}
\toprule
\bf Hyperparameters & \bf L=12 & \bf L=24 \\
\midrule
Peak learning rate & 5e-4 & 3e-4\\
Fine-tuning epochs & 100  & 50 \\
Warmup epochs & 20 & 5 \\
Layer-wise learning rate decay & 0.65 & 0.8 \\
Batch size & \multicolumn{2}{c}{1024} \\
Adam $\epsilon$ & \multicolumn{2}{c}{1e-8}  \\
Adam $\beta$ & \multicolumn{2}{c}{(0.9, 0.999)} \\
Minimal learning rate & \multicolumn{2}{c}{1e-6} \\
Learning rate schedule & \multicolumn{2}{c}{Cosine} \\
\midrule
Repeated Aug & \multicolumn{2}{c}{\xmark} \\
Weight decay & \multicolumn{2}{c}{0.05} \\
Label smoothing $\varepsilon$ & \multicolumn{2}{c}{0.1}     \\
Drop path & 0.1 & 0.2 \\
Dropout & \multicolumn{2}{c}{\xmark} \\
Gradient clipping & \multicolumn{2}{c}{\xmark} \\
\midrule
Erasing prob.  & \multicolumn{2}{c}{0.25} \\
Input resolution & \multicolumn{2}{c}{$224 \times 224$} \\
Rand Augment  & \multicolumn{2}{c}{9/0.5} \\
Mixup prob.  & \multicolumn{2}{c}{0.8}     \\
Cutmix prob.   & \multicolumn{2}{c}{1.0}    \\
\bottomrule
\end{tabular}
}
\caption{
Hyperparameters for fine-tuning \our{} on ImageNet-1K.
}
\label{tbl:ft:imagenet:hyperparams}
\end{table}

\begin{table}[ht]
\centering
\scalebox{0.95}{
\begin{tabular}{l|c|c}
\toprule
\bf Hyperparameters & \bf L=18   & \bf L=36\\
\midrule
Layers & 18 & 36 \\
Hidden size & 512 & 512 \\
FFN inner hidden size & 2048 & 2048\\
Attention heads & 8 & 8\\
Relative positional embeddings & \checkmark & \checkmark \\
\midrule
Training steps & 400K & 400K \\
Epochs & 150 & 150 \\
AdamW $\epsilon$ & 1e-6& 1e-6 \\
AdamW $\beta$ & (0.9, 0.98)& (0.9, 0.98) \\
Peak learning rate & 5e-3  & 3e-3 \\
Learning rate schedule & Linear& Linear \\
Warmup steps & 32k & 32k\\
\midrule
Gradient clipping & 1.0 & 1.0 \\
Dropout & 0.1 & 0.1\\
Weight decay & 0.01 & 0.01\\
\midrule
Speed perturbation & \xmark & \xmark\\
Frequency masks &  2  &  2  \\
Maximum frequency-mask width &  27 &  27 \\
Time masks & 10 & 10  \\
Maximum time-mask ratio &0.04&0.04  \\

\bottomrule
\end{tabular}
}
\caption{
Hyperparameters for training \our{} on LibriSpeech.
}
\label{tbl:TT:hyperparams}
\end{table}

\begin{table}[ht]
\centering
\scalebox{0.95}{
\begin{tabular}{l|c}
\toprule
\bf Hyperparameters & \bf BEiT-3 pretraining \\
\midrule
Layers & 24 \\
Hidden size & 544 \\
FFN inner hidden size & 2176 \\
Attention heads & 16 \\
Patch size & $16 \times 16$ \\
Relative positional embeddings & \xmark \\
\midrule
Training steps & 300K \\
Batch size & 12288 \\
AdamW $\epsilon$ & 1e-6 \\
AdamW $\beta$ & (0.9, 0.98) \\
Peak learning rate & 2.8e-3 \\
Learning rate schedule & Cosine \\
Warmup steps & 20k \\
\midrule
Gradient clipping & 3.0 \\
Dropout & \xmark \\
Drop path & 0.1 \\
Weight decay & 0.05 \\
\midrule
Data Augment & RandomResizeAndCrop \\
Input resolution & $224^2$ \\
Color jitter & 0.4 \\
\bottomrule
\end{tabular}
}
\caption{
Hyperparameters for vision-language pretraining.
}
\label{tbl:pretrain:hyperparams}
\end{table}

\begin{table}[ht]
\centering
\scalebox{0.95}{
\begin{tabular}{l|cc}
\toprule
\bf Hyperparameters & \bf NLVR2 & \bf VQA \\
\midrule
Peak learning rate & \multicolumn{2}{c}{\{1e-5, 2e-5, 3e-5\}} \\
Fine-tuning epochs & \multicolumn{2}{c}{10} \\
Warmup epochs & \multicolumn{2}{c}{1} \\
Layer-wise learning rate decay & \multicolumn{2}{c}{1.0} \\
Batch size & \multicolumn{2}{c}{128} \\
AdamW $\epsilon$ & \multicolumn{2}{c}{1e-8}  \\
AdamW $\beta$ & \multicolumn{2}{c}{(0.9, 0.999)} \\
Weight decay & \multicolumn{2}{c}{0.01} \\
Drop path & 0.2 & 0.1 \\
Dropout & \multicolumn{2}{c}{\xmark} \\
Input resolution & $224^2$ & $384^2$ \\
\bottomrule
\end{tabular}
}
\caption{
Hyperparameters for fine-tuning \our{} and the baseline on NLVR2 and VQA.
}
\label{tbl:ft:vqa_nlvr2:hyperparams}
\end{table}

\end{document}